# Material-Segmented Per-Pixel Emissivity Correction for Thermographic Anomaly Detection in Cultural Heritage Digital Twins

## 面向文化遗产数字孪生中热成像异常检测的材料分割逐像素发射率校正

Jonathan Klingspon[1], Scott McAvoy[1], Maurizio Seracini[2], Falko Kuester[1]

[1] University of California, San Diego — J. Klingspon (Ph.D. Student, Computer Engineering, Dept. of Computer Science and Engineering); S. McAvoy (Data Scientist); F. Kuester (Professor; Director and P.I., Cultural Heritage Engineering Initiative). [2] San Diego State University — M. Seracini (Professor).

Corresponding author: Jonathan Klingspon, jklingsp AT ucsd.edu

**Abstract:** Quantitative longwave thermography of heritage surfaces is limited by the global-constant emissivity assumption in inverse-Planck temperature retrieval; on heterogeneous surfaces emissivity varies within one field of view, producing apparent-temperature artifacts that mimic and mask subsurface anomalies. We present a training-free pipeline that derives per-pixel emissivity by applying SAM 3.1 open-vocabulary segmentation to a colocated, co-calibrated RGB channel, mapping segments to a material-keyed LWIR emissivity table compiled from primary measurement literature, and propagating the field into a per-pixel inverse-Planck solve on raw radiometric data. Lacking any public dataset with raw radiometry, a temperature reference, and a colocated RGB camera, we evaluate on a physics-based synthetic benchmark and four real datasets. On the benchmark, under a palette spanning the low-emissivity exceptions, the correction cuts mean absolute error from 1.97 K to 0.91 K at 20 K contrast and, with an accurate table, beats the best fitted global constant on every layout; on a heritage-realistic emissivity distribution it does not. We contribute a quantified operating-regime map, and a measurement-backed finding that tempers the heritage claim: weathered outdoor heritage emissivities cluster near the conventional default, so the correction is small on typical surfaces and concentrated on genuine low-emissivity exceptions. We characterize the dominant failure mode, in which open-vocabulary segmentation matches appearance rather than material, and the contraindicated regime in which emissivity-defined anomalies are suppressed.

**摘要：**遗产表面的定量长波热成像受制于反演普朗克测温中的全局恒定发射率假设。在非均质表面上，发射率在同一视场内变化，会产生模拟并掩盖次表层异常的表观温度伪影。本文提出一种免训练流程：将 SAM 3.1 开放词汇分割（open-vocabulary segmentation）应用于并置且共标定的 RGB 通道，把分割区域映射到依据一手测量文献编制的材料–长波红外发射率查找表，并将该逐像素发射率场代入对原始辐射数据的逐像素反演普朗克求解。由于不存在同时提供原始辐射、独立温度基准与并置 RGB 相机的公开数据集，我们在一个基于物理的合成基准和四个真实数据集上进行评估。在合成基准上，当调色板涵盖低发射率特例时，该校正在 20 K 对比度下将平均绝对误差由 1.97 K 降至 0.91 K，且在表格准确时于每个布局上均优于最佳拟合全局常数；但在符合遗产实际的发射率分布下则不然。我们给出量化的适用区间图，以及一个由测量支撑、对遗产主张予以收敛的结论：风化室外遗产材料的发射率聚集于常规默认值附近，故在典型表面上校正很小，仅集中于真正的低发射率特例。我们刻画了主要失效模式——开放词汇分割匹配外观而非材料——以及由发射率对比定义的异常会被抑制的禁用区间。

## 1. Introduction

Thermal imaging sensors do not measure temperature directly, but rather the incident infrared radiance at the sensor, which is dominated by surface emission and reflection together with emission and attenuation from the intervening atmosphere. For a gray surface the total radiance is $W_tot = \varepsilon \cdot \tau \cdot W_obj + (1-\varepsilon) \cdot \tau \cdot W_refl + (1-\tau) \cdot W_atm$, where W_obj is the radiance from an object at its true temperature, W_refl is the radiance reflected from the surrounding apparent temperatures, W_atm is the atmospheric radiance, $\varepsilon$ is the surface emissivity, and $\tau$ is the atmospheric transmissivity. Recovering the object temperature therefore requires $\varepsilon$ and W_refl to be known, and operators almost universally supply a single scalar emissivity for the whole image; the FLIR application records a preset 0.95 in every frame header. Two surfaces at the same temperature may render differently in a thermograph depending on their emissivity, and while this can be corrected for known materials using emissivity tables, doing so is a time-intensive process requiring manual effort and prior knowledge. On a homogeneous, high-emissivity surface a single value is adequate, but heritage surfaces are neither homogeneous nor uniformly high: marble, plaster, patinated bronze, gilding, wood, glass, and bare metal can occur in one field of view with emissivities from below 0.1 to above 0.98. Under a global-constant assumption each material whose true emissivity departs from the assumed value is assigned a biased temperature, and the bias appears as a spatial discontinuity aligned with the material boundary rather than with any real thermal feature. Such artifacts both mimic and mask the subsurface anomalies, delaminations, moisture fronts, and voids, that quantitative thermography is deployed to find.

The problem is underdetermined. A single band provides one measurement for two unknowns plus the reflected term. The multiband temperature and emissivity separation used in remote sensing, such as the ASTER algorithm, closes the system with an empirical spectral constraint across five thermal bands, and even then reaches only ±1.5 K and ±0.015 in emissivity[1]. A commodity single-band imager has no such spectral leverage. Our premise is that a colocated visible-light camera supplies the missing channel: the material identity that is visually obvious in the RGB image constrains the emissivity that is invisible in the thermal image. Recent open-vocabulary segmentation models make this practical without task-specific training, because the materials of a scene can be named directly as free-text prompts.

This work makes four contributions. First, we implement and validate a training-free pipeline that segments a colocated RGB image with open-vocabulary prompts, maps the resulting material regions to a curated LWIR emissivity lookup table, and applies the per-pixel emissivity field in a complete raw-digital-number inverse-Planck retrieval on commodity and professional hardware (Fig. 1), demonstrated in situ on the Loggia dei Lanzi. Second, because no public dataset provides raw radiometric data, an independent temperature reference, and a colocated calibrated RGB camera at once, we construct a physics-based synthetic benchmark that is exactly invertible and use it to measure the accuracy improvement against known truth. Third, we contribute a quantified operating-regime map: rather than claim a uniform improvement, we characterize the thermal contrast, emissivity diversity, lookup-table accuracy, registration, and calibration conditions under which correction helps, is inert, or does harm. Fourth, we compile a primary-source LWIR emissivity table for the specific weathered heritage materials of our scenes and use it to state a tempering result: the measured emissivities of weathered outdoor heritage surfaces cluster near the conventional default, so the correction is small there and concentrated on the low-emissivity exceptions. The core mechanism of visible-image segmentation driving per-material emissivity correction is not new; our specific instantiation, combining training-free open-vocabulary segmentation, a full raw-DN radiometric chain, in-situ heritage deployment, the contraindication analysis, and the measurement-backed emissivity audit, is to our knowledge without precedent.

## 2. Related Work

Visible-image emissivity correction predates the deep-learning era. Ding et al. used a binocular visible-plus-infrared rig with Canny edge extraction, manually seeded region growing, and per-region emissivity assignment; the method was region-level and non-semantic, with no temperature ground truth[2]. Gao and Tian correlated registered visible and infrared spectra to correct emissivity per pixel for non-destructive evaluation, again without material semantics[3]. The closest prior art is Schmid et al., who trained a closed-set convolutional network to segment a visible image of a laboratory specimen into material classes, measured each class emissivity by oven heating and contact thermometry, registered the frames manually, and corrected the thermogram per class[4]. Their evaluation is confined to a single multi-material plate and a printed circuit board under laboratory heating, and their pipeline is explicitly closed-set: it applies only when the network has been trained on the exact materials later imaged. A direct comparison is infeasible because that fixed material set, the unavailable imagery, and the contact protocol have no field-radiometric analogue, so we position the two methods by design rather than a shared benchmark. Our method differs in requiring no task-specific training data, in operating on raw radiometric data in the field, and in supplying a quantified account of when the correction should and should not be trusted.

Where our approach substitutes a visual prior for spectral information, the heat-assisted detection and ranging (HADAR) framework resolves the temperature and emissivity ambiguity directly using hyperspectral longwave imaging and a learned decomposition into temperature, emissivity, and texture[5]. HADAR reaches near-optimal accuracy but requires spectral hardware; we position the RGB semantic prior as a low-cost substitute on a single-band imager, and we reuse the HADAR emissivity library as an external cross-check for our lookup table.

Quantitative infrared thermography of cultural heritage is well established[6,7], and the sensitivity of retrieved temperature to emissivity is a recognized error source; standard practice determines emissivity in the laboratory on representative samples and applies it per material or region, with no automated material recognition. Per-region emissivity and reflected-energy correction has been applied to as-is three-dimensional thermal models of buildings[8], and the resulting material and temperature information increasingly feeds heritage digital twins[9] and public-facing augmented-reality experiences that overlay diagnostic imaging on monuments[10], which motivates delivering the emissivity and temperature fields as separately addressable channels. The radiometric measurement chain is standard[11], and reference practice for establishing reflected temperature is codified in ASTM E1862[12].

## 3. Materials and Methods

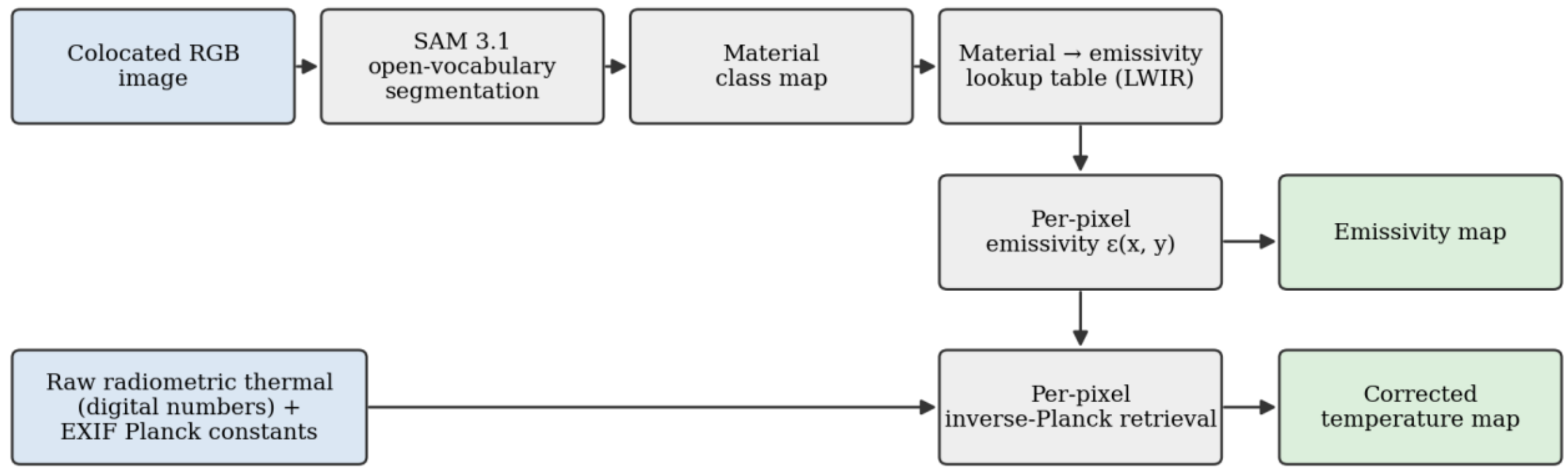

*Figure 1. Overview of the method. Open-vocabulary segmentation of the colocated RGB image assigns each pixel a material, a lookup table maps materials to LWIR emissivity, and the resulting per-pixel emissivity field drives an inverse-Planck retrieval of the colocated raw radiometric data, producing coregistered corrected-temperature and emissivity rasters.*

### 3.1 Radiometric model

The raw outputs of the FLIR One Pro are JPEG images whose EXIF headers embed the raw thermal sensor data, the embedded RGB image of the colocated camera, the manufacturer-defined per-sensor Planck calibration constants (R1, R2, B, F, O), and the user-set atmospheric and capture parameters. We recover these fields from the headers and apply the FLIR Planck and digital-number parameterization[11,13]. The signal of a blackbody at temperature T is $S(T) = R1/(R2(\exp(B/T) - F)) - O$, and its exact inverse, which maps the corrected object signal to temperature in degrees Celsius, is $T(S) = B/\ln(R1/(R2(S+O)) + F) - 273.15$. The measured signal composites object, reflected, atmospheric, and IR-window contributions: the object and reflected terms are attenuated by the object-path gain $G = \tau 1 \cdot \iota \cdot \tau 2$, and two atmospheric legs over half the object distance and an IR-window term of transmission $\iota$ and temperature T_wind are added. Given an emissivity field the composite is solved per pixel for the object signal and inverted; retrievals leaving the physical domain return not-a-number rather than a spurious finite temperature. The IR-window term is not optional for the FLIR One Pro, whose header carries a window transmission of 0.80: omitting it biases retrieval by +2.1 K at 15 °C and −2.9 K at 40 °C. Our implementation reproduces the reference to within 1e-9 K, and forward and inverse models are exact algebraic inverses to within 1e-8 K, so any error in the results is attributable to the emissivity model and sensor degradations, not the radiometric code.

### 3.2 Emissivity sensitivity and the degeneracy

The temperature error induced by an emissivity error follows from differentiating the inverse chain. Writing A for the emissivity-independent part of the object signal, $\partial T/\partial \varepsilon \propto (S_refl - A)/\varepsilon^2$. Two properties govern the method. First, on a physically consistent scene the reflected minus object gap is itself proportional to emissivity, $S_refl - A = \varepsilon(S_refl - S_obj)$, so one power cancels and both emissivity-lookup error and additive sensor noise inflate the retrieved-temperature error as $1/\varepsilon$ as emissivity falls. Low-emissivity surfaces are therefore where the correction matters most and is least reliable. Second, the sensitivity vanishes as T_obj approaches T_refl: when the object is at the reflected apparent temperature the measured radiance is independent of emissivity, and no correction is observable. This degeneracy is exact, so a scene lacking thermal contrast cannot demonstrate or refute an emissivity method.

### 3.3 Open-vocabulary material segmentation

Material regions are obtained from the colocated RGB image with SAM 3.1 promptable-concept segmentation[14], the open-vocabulary successor to the Segment Anything family[15]. Each material is supplied as a free-text prompt, for example bronze statue, marble sculpture, or glass window; the model grounds each prompt directly and returns instance masks with confidence scores, with no task-specific training. Masks from all prompts are composited into a class map by a per-pixel score-argmax rule, with materials ordered so that ties prefer the higher-emissivity, lower-correction label. Unmatched pixels retain the frame default emissivity, so unclassified regions are left uncorrected rather than miscorrected. Segmentation quality, not the correction physics, is the practical bottleneck; the method depends on RGB quality, and the relevant metric is emissivity-weighted confusion, since a marble and plaster confusion is harmless at $\Delta\varepsilon \approx 0.02$ whereas a marble and gilding confusion spans $\Delta\varepsilon \approx 0.9$ and drives a double-digit error through the $1/\varepsilon$ amplification.

### 3.4 Emissivity lookup table

Material classes map to LWIR emissivities through a curated table assembled from primary measurement literature where available and consensus technical tables otherwise, with source and uncertainty recorded per entry (Table 1). Three findings shaped it. First, there is no published direct LWIR measurement of named pietra serena or pietra forte, so we use Mineo and Pappalardo's measured sandstone and limestone as mineralogical proxies[16], informed by the spectral structure of terrestrial silicates and carbonates in the 8 to 14 μm window[17]. Second, weathered outdoor architectural metal is high-emissivity, at 0.85 to 0.98[18,19], not the polished-metal 0.15 to 0.35[20], and the polished value applied to weathered metal produces spurious double-digit corrections through the 1/ε amplification. Third, the measured emissivities of realistic weathered heritage surfaces cluster near the 0.95 default: patinated bronze about 0.88[21,22,23], Carrara marble about 0.96[24], lime plaster about 0.94[25], weathered wood about 0.92[26], and human skin 0.98[27]. As an external check, band-averaging the HADAR spectral emissivity library[5] reproduces our marble, stone, and plaster entries to within 0.05, but its remaining materials have no clean counterpart to ours, since its only glass is a low-emissivity crystalline glass rather than an opaque soda-lime pane; the check therefore covers the common silicate and carbonate classes rather than the full table, and the low-emissivity entries that dominate the correction remain externally unvalidated.

| Material (weathered outdoor) | ε | Uncertainty | Basis / confidence |
|---|---|---|---|
| Patinated bronze (statue) | 0.88 | ±0.08 | FTIR oxidized-Cu plus IRT practice (weak) |
| Carrara marble, weathered | 0.96 | ±0.02 | primary-measured (well-grounded) |
| Pietra serena (sandstone) | 0.94 | ±0.03 | sandstone proxy (weak) |
| Pietra forte / limestone | 0.96 | ±0.02 | calcarenite/limestone proxy (moderate) |
| Lime plaster / render | 0.94 | ±0.03 | mortar 0.92 plus weathering (well-grounded) |
| Interior gypsum plaster | 0.82 | ±0.03 | a genuine low-ε case (well-grounded) |
| Weathered wood (shutters, doors) | 0.92 | ±0.04 | pooled species plus weathering |
| Soda-lime glass, near-normal | 0.91 | ±0.03 | technical tables; dips at grazing incidence |
| Terracotta / fired brick | 0.94 | ±0.02 | weathered fired clay, 8 to 14 μm |
| Weathered architectural metal | 0.94 | ±0.06 | oxidized iron/magnetite (moderate) |
| Bare / polished metal (reference) | 0.15 | ±0.10 | do not use unless truly polished |
| Human skin | 0.98 | ±0.01 | primary-measured (well-grounded) |

*Table 1. Recommended LWIR (8 to 14 μm) emissivity for the heritage materials of the study, favouring primary-measured values for realistic weathered outdoor finishes.*

### 3.5 RGB and thermal registration

The emissivity map computed in the RGB frame must be transferred to the thermal grid. As established for the FLIR One Pro by McAvoy et al.[13], the camera alignment metadata encodes only a scale-and-translate model (an X and Y pixel offset and an RGB-to-IR scale factor) that ignores lens distortion and parallax and fails most visibly near the image edges. For the FLIR One Pro pulpit set the available planar calibration is geometrically degenerate, so we transfer the map with a single global homography, which for a roughly 1 cm baseline at 1.5 to 5 m incurs a residual parallax below a few thermal pixels. For the FLIR T1020 the RGB and thermal cameras are rigidly colocated in one body, but their alignment is depth-dependent, so a single fixed homography is inappropriate. We recover a fixed relative pose from a field-improvised heated target of 48 metrically measured corners, with the intrinsics fixed to their planar calibrations[28]; the stereo solve yields a physically sensible 6.5 cm baseline and a whole-target reprojection residual of about 4.5 pixels. The per-frame alignment is a plane-induced homography at the EXIF focus distance. Because a single plane aligns only surfaces at the warp plane, a surface at depth z incurs parallax $f \cdot b \cdot (1/z - 1/z_{focus})$, which for f = 2137 pixels and b = 6.5 cm reaches about 40 pixels across a 3 to 20 m spread, so the warp is accurate only within a narrow depth band and the foreground Perseus is treated qualitatively. A multi-camera structure-from-motion export of the same set produced a physically impossible 0.49 to 1.55 m baseline, so the constrained target solve is required. Where a plane-free warp is needed, learned cross-modal matchers and monocular metric depth provide dense correspondences and per-pixel depth; we did not require them for the near-planar frames analysed quantitatively.

### 3.6 Baselines and statistical treatment

Every experiment compares four emissivity assignments on identical inputs: the constant frame default $\varepsilon = 0.95$ (B0), the standard workflow the method must beat; the single scalar $\varepsilon$ that best fits the ground truth (B1), separating a better scalar from per-pixel structure; the material-informed map (M); and the true per-pixel emissivity (O), an oracle upper bound isolating segmentation and lookup-table error from the retrieval physics. The unit of replication in the synthetic benchmark is the material layout, not the pixel. The eight layouts are real class-map geometries from one site sharing a fixed emissivity palette, so between-layout variance reflects geometry rather than independent material draws; confidence intervals are Student-t intervals over the eight layouts. On real data, intervals are cluster bootstraps over frames, except the ULB17 correlation, which is reported descriptively. The primary endpoint is the paired per-layout mean absolute error improvement of M at 20 K contrast with an accurate table, reported against B0 and, as the co-primary comparison at an accurate table, against B1.

### 3.7 Synthetic ground-truth benchmark

Because no real dataset provides per-pixel temperature ground truth together with a colocated RGB camera, the accuracy claim rests on a physics-based synthetic benchmark that is exactly invertible by construction. Scenes are generated on the native 160×120 grid of the sensor microbolometer core, using real material layouts and real per-frame calibration constants. A ground-truth temperature field, with a smooth ambient component, material-correlated offsets, and injected anomalies some straddling material boundaries, and a ground-truth emissivity field are forward-rendered to raw digital numbers through the full chain. Emissivities are assigned under one of two palettes: a diverse palette [0.94, 0.78, 0.92, 0.40, 0.90, 0.88, 0.85, 0.10] that deliberately includes low-emissivity materials to stress-test the method on the genuine low-$\varepsilon$ exceptions rather than on typical high-$\varepsilon$ heritage surfaces, and a heritage-realistic site palette of 0.88 to 0.95 taken from the measured pulpit table. A sensor degradation stack, with optical blur, temporal noise at 60 mK, fixed-pattern residual non-uniformity, and radial falloff[29], is applied in digital-number space and referred through the detector gain at unit emissivity, so the emissivity-dependent noise amplification is measured rather than assumed, followed by the interpolated up-sampling that mimics the camera. To break the inverse crime, generation perturbs the Planck constants R1, O, and B at the level of plausible factory-calibration error while inversion uses nominal constants. On this benchmark M applies the lookup table to the ground-truth class map, so true material identity is assumed and the accurate-table, zero-corruption figures upper-bound the deployed segmentation pipeline; segmentation and registration error are injected separately as controlled sweeps.

### 3.8 Datasets

We evaluate on four real datasets chosen to span the operating regime. The pulpit set is 207 raw radiometric frames of the Brunelleschi pulpit in Santa Maria Novella (FLIR One Pro), an indoor, near-isothermal, predominantly high-emissivity site. ULB17-VT provides 404 natively-aligned RGB and radiometric per-pixel-temperature pairs (FLIR E60), outdoor scenes with large thermal contrast reaching 71 °C within a frame and diverse materials[30]. ThermoScenes provides aligned RGB and thermal building facade scenes (FLIR One Pro LT)[31]. The Loggia dei Lanzi set (FLIR T1020) provides native 1024×768 radiometric frames with a colocated 1280×960 RGB camera and a ±1 °C accuracy class, comprising outdoor facades and the Loggia dei Lanzi statues including Cellini's Perseus, acquired during the Florence thermography and XR field campaign of December 2025 and January 2026[32]. None provides independent contact temperature ground truth, so on real data we report relative and behavioural metrics.

## 4. Results

### 4.1 Accuracy against synthetic ground truth

On the diverse palette at 20 K contrast, with the full noise stack and a 0.5% calibration mismatch, whole-frame mean absolute error falls from 1.97 K (B0) to 0.91 K (M) (Table 2), near the oracle 0.90 K (O); interior error, excluding the material-boundary band, falls from 1.79 to 0.65 K. The paired improvement of M over B0 is +1.06 K (95% CI [0.58, 1.54], all eight layouts). M also beats the best fitted constant B1 (1.78 K) by +0.87 K ([0.50, 1.25], all eight) at an accurate table, but this margin falls to +0.14 K ([−0.40, 0.69]) at a per-class table error of 0.05, the tolerance of the HADAR check. One illustrative layout falls from a 2.45 K constant-default error to 0.73 K toward its oracle 0.55 K, the error collapsing from an area-wide bias on the low-emissivity regions to thin residual lines at material boundaries (Fig. 2). The advantage is contrast-dependent, since at 10 K it is +0.46 K over B0 with no separation from B1, and it is palette-dependent: on the heritage-realistic site palette M improves over B0 by only +0.07 K and is worse than the best scalar, at −0.28 K. On a realistic weathered-heritage distribution the per-pixel map does not beat a well-chosen constant; its value is confined to scenes that contain the low-emissivity exceptions, exactly as the heritage audit and the regime map predict.

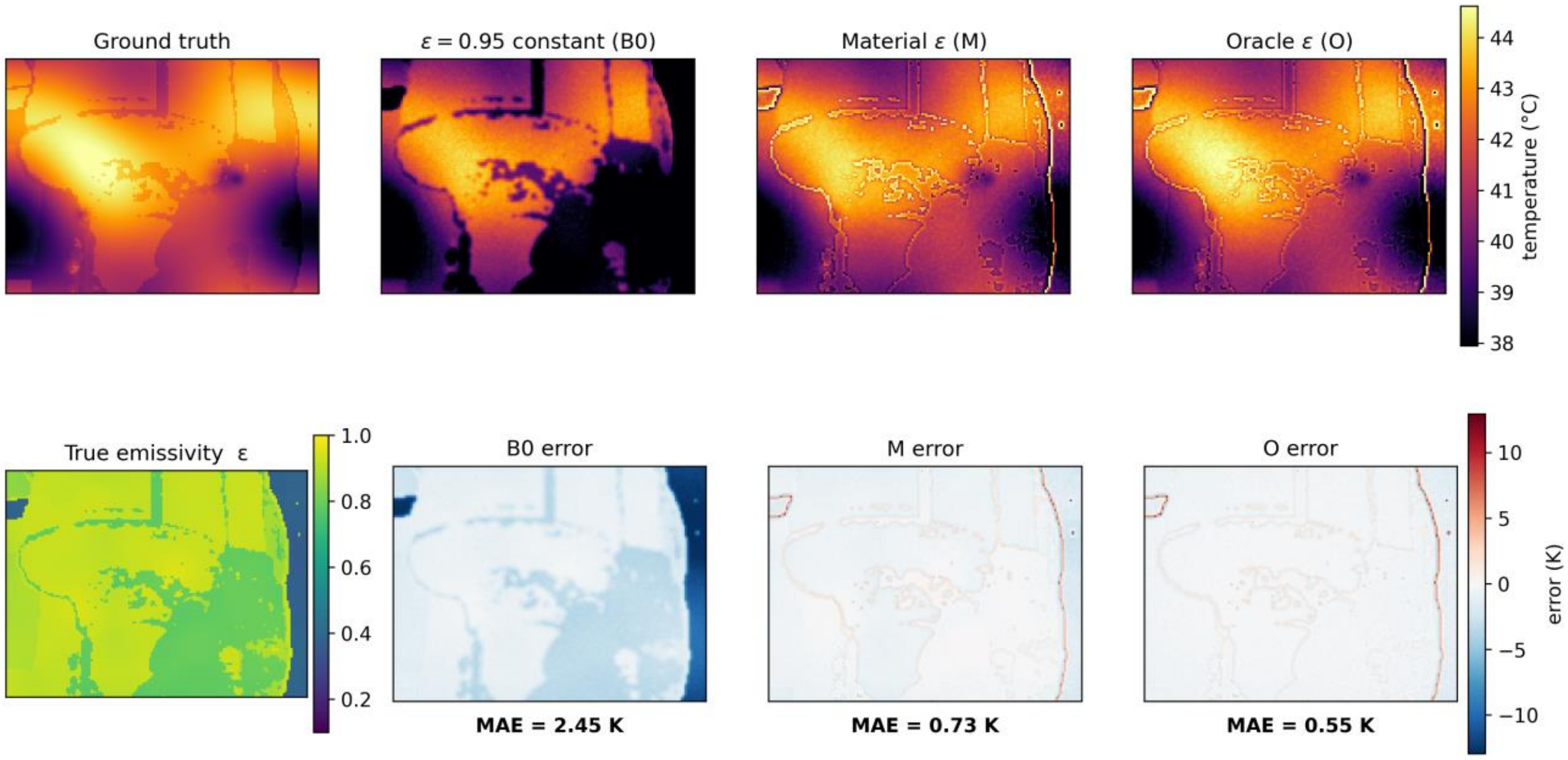

*Figure 2. Synthetic benchmark on one illustrative layout at 20 K contrast (mean absolute error 2.45 to 0.73 K; the eight-layout mean is 1.97 to 0.91 K). Top row: ground truth and the three retrievals. Bottom row: the true emissivity field, then the per-method error against ground truth (B0, M, O).*

| Method | MAE at 10 K (K) | MAE at 20 K (K) |
|---|---|---|
| B0, constant default ε = 0.95 | 1.21 | 1.97 |
| B1, best fitted scalar | 0.93 | 1.78 |
| M, material-informed map | 0.76 | 0.91 |
| O, oracle per-pixel truth | 0.76 | 0.90 |

*Table 2. Whole-frame mean absolute error by method and scene contrast, aggregated over eight layouts (diverse palette, full noise, 0.5% mismatch). Paired M over B0: +0.46 K at 10 K, +1.06 K at 20 K; M over B1: +0.17 K (not significant) at 10 K, +0.87 K at 20 K.*

### 4.2 Operating-regime map and sensitivity sweeps

The improvement grows with contrast and shrinks with table error (Fig. 3). At 20 K it is +1.06 K with an accurate table, +0.33 K at 0.05 table error, and net harm at −0.52 K by 0.10, and the crossover to harm occurs at smaller table error as contrast falls. Corruption sweeps set concrete tolerances. At 10 K, where the improvement over B0 is +0.46 K, segmentation errors erode it to net harm by a 20% class-flip rate, misregistration erodes it to break-even by about 8 thermal pixels, and a ±5 K reflected-temperature error erases it. These tolerances are the quantitative basis for treating segmentation as the practical bottleneck, and the 8-pixel registration budget is exceeded by the Loggia's roughly 40-pixel parallax, which is why the T1020 numbers are confined to near-planar facade frames. A separate sensitivity is to calibration itself: a 1.5% error in the device Planck constants inflates the retrieval error until the benefit collapses to within noise, so absolute-accuracy claims require a trustworthy calibration and on uncalibrated commodity hardware the defensible claims are relative and spatial.

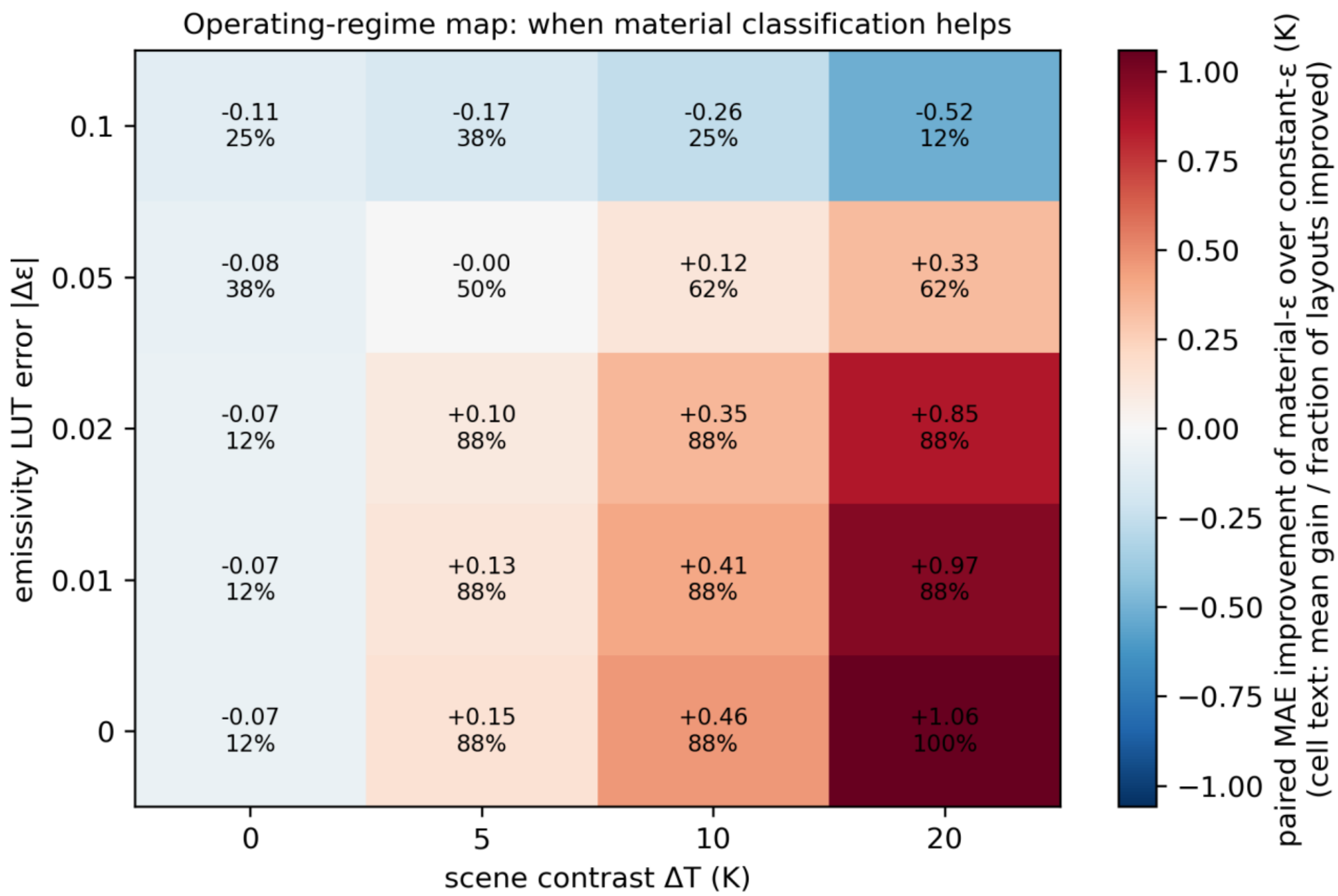

*Figure 3. Operating-regime map. Paired mean absolute error improvement of the material map over the constant-default baseline, as a function of scene thermal contrast and per-class lookup-table error, over eight layouts.*

### 4.3 Behavioural validation on real high-contrast data

On ULB17-VT, whose retrieved temperature we re-solve through the two-term model under a boxcar 8 to 14 µm response, the correction magnitude increases with the deviation of the material emissivity from the default (Spearman $\rho = 0.65$, $n = 38$ material instances). This is a consistency check, not an independent-accuracy test: the correction is derived from the same lookup-table emissivity it is correlated against, so a positive monotone relationship is expected by construction, and we claim no p-value or interval because the instances are not independent and six frames are too few clusters for a trustworthy bootstrap. Near-blackbody vegetation and skin are barely altered, at a mean 0.24 K, which serves as a control. Glass and wood shift by +1.4 K and +1.3 K, and propagating the lookup-table emissivity uncertainty through the sensitivity relation attaches 1σ intervals of ±0.5 K for glass, which is well resolved, and ±1.0 K for wood, which is marginal; the relevant yardstick is this propagated uncertainty, not the 60 mK sensor noise. The largest single-instance value, +23.6 K on a region labelled metal pole, is a segmentation and lookup-table artifact rather than a genuine bare-metal correction: it is weathered wood mislabelled as metal at ε = 0.35, inflated by the 1/ε amplification, and it is numerically indistinguishable from a correction one would expect on real bare metal, which is why such low-emissivity corrections must carry propagated uncertainty rather than a point value.

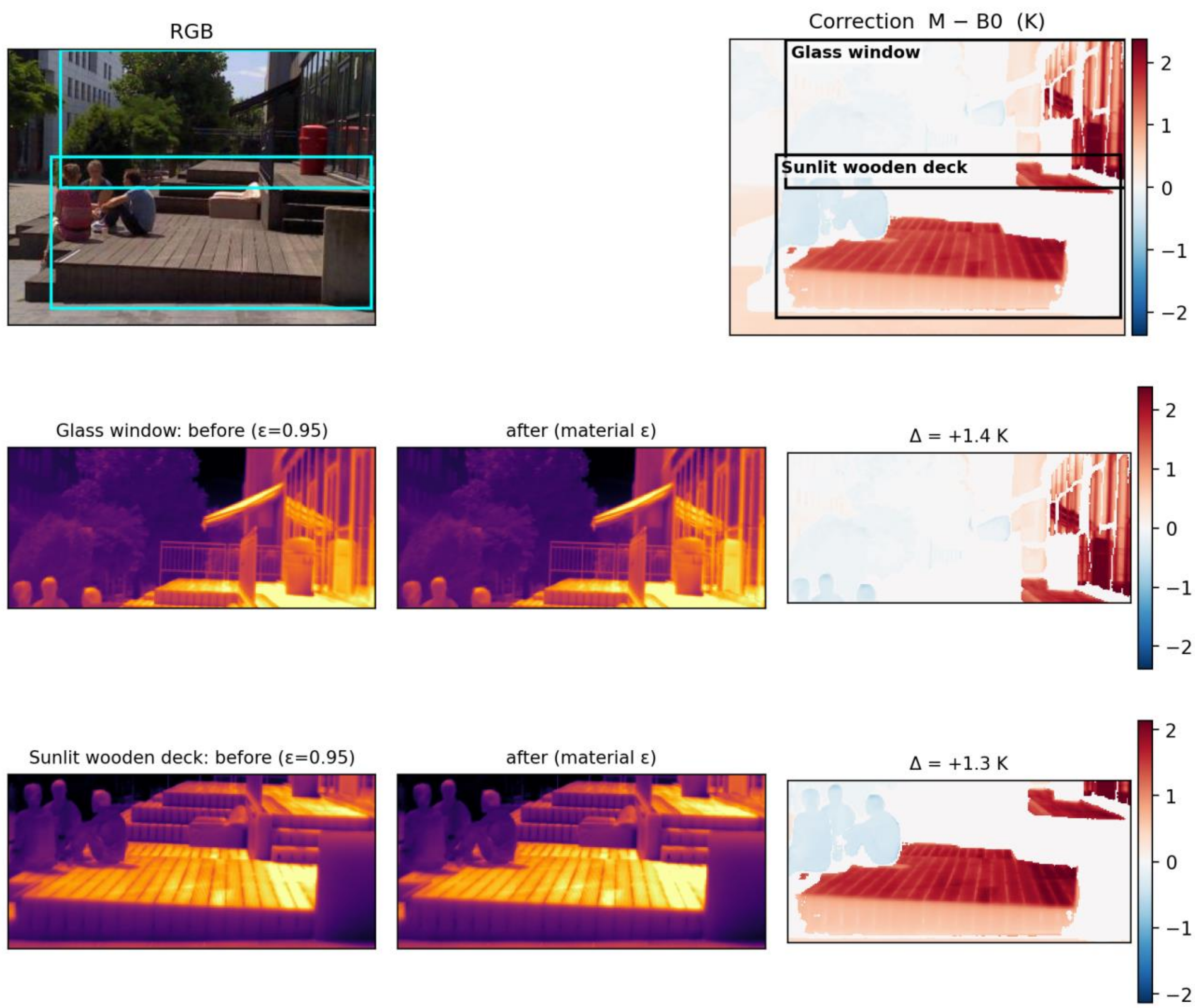

*Figure 4. ULB17-VT frame 200. Top: RGB and the full-frame correction map (M − B0), with the glass-window and sunlit-deck regions boxed. Below, for each region: the constant-emissivity (B0) and material-emissivity (M) retrievals under a shared temperature scale and their difference, zoomed to the region.*

### 4.4 Heritage deployment: the Loggia dei Lanzi

On the well-lit arcade, SAM grounds Perseus at instance level as bronze and separates the surrounding marble statues; the architecture is labelled by visual appearance rather than verified material identity, which is harmless because those classes are all high-emissivity at 0.94 to 0.96. With the measurement-backed emissivities the correction on the Loggia scenes is small (whole-frame 95th percentile about 0.1 K) and concentrated on the Perseus bronze, where it reaches about 0.7 K (the cold statue reads cooler than the constant-ε default because it sits well below the reflected field). Under the EXIF reflected estimate this is not a low-contrast artifact: the surface sits about 9 K below the reflected apparent temperature (a roughly 9 °C scene against an 18 °C reflected field), nominally placing the retrieval away from the emissivity degeneracy. That gap, however, inherits the reflected-temperature uncertainty quantified below, since a cold-sky departure of 5 to 10 K would collapse much of it, so we read the small correction as consistent with, rather than proof of, small emissivity deviations. Three caveats bound the bronze number. The reflected apparent temperature is held as a single scalar (EXIF 18 °C) and was not measured, whereas the open arcade has a strongly directional reflected field; at $(1 - \varepsilon)/\varepsilon = 0.14$ K per K a 5 to 10 K reflected-field error injects 0.7 to 1.4 K, comparable to the correction. The tabulated bronze uncertainty of ±0.08 attaches a further ±0.7 K, so the bronze correction is a weak-confidence result not statistically separable from zero. And Perseus, as a foreground object in a receding arcade, has a mask boundary uncertain to the foreground parallax. Inspecting four arcade frames against the RGB confirms that SAM makes no emissivity-costly error: the one emissivity-distinct material, bronze, is correctly isolated with no false bronze on the marble statues, so the emissivity error induced by segmentation on the arcade is negligible. On the outdoor facade session, captured at night, the RGB is dark and the classified fraction collapses to 0 to 8%, so RGB quality, not the correction physics, is the practical limiter and the method inherits the visible camera operating envelope (Fig. 5).

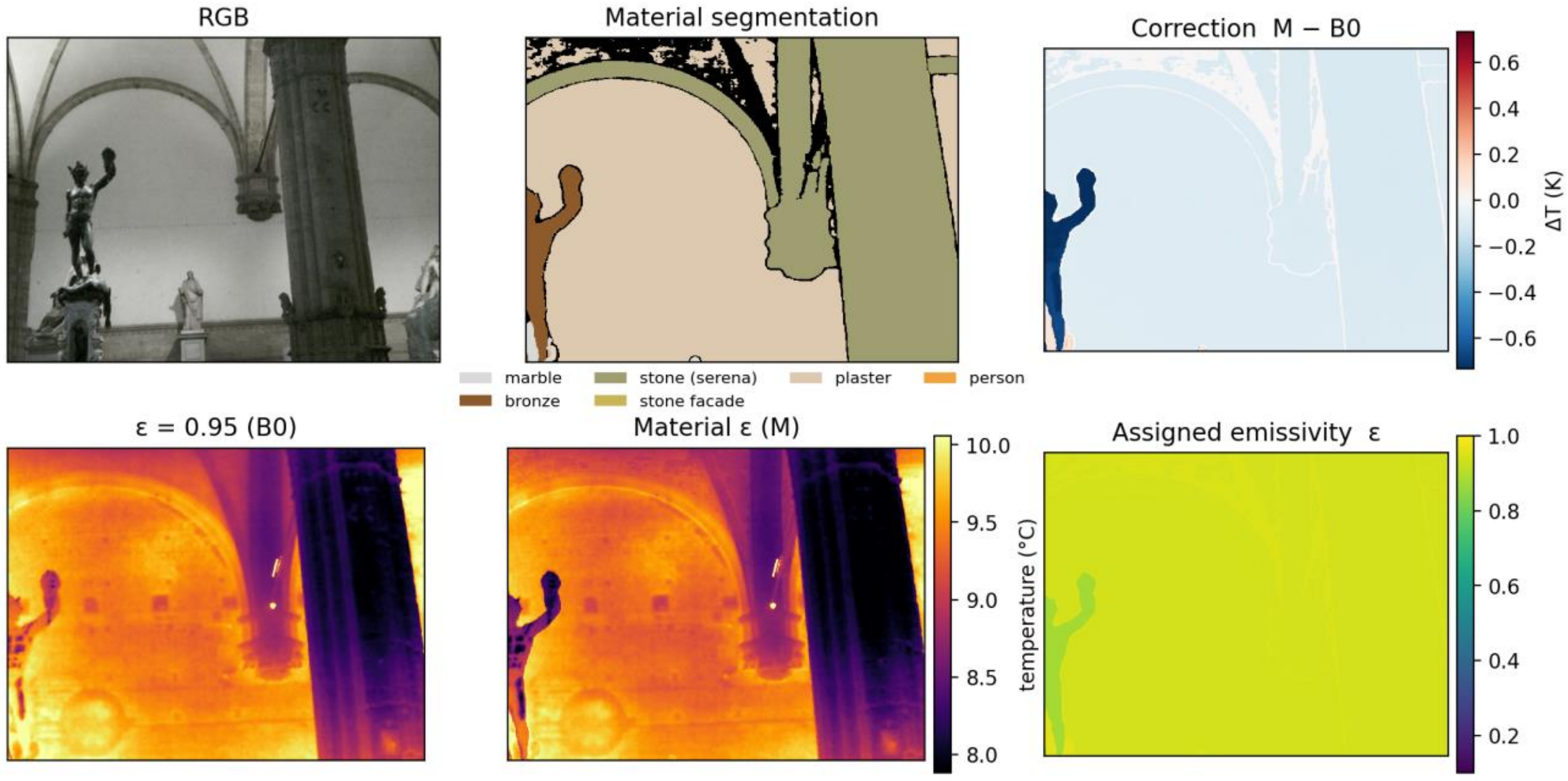

*Figure 5. Full pipeline on the Loggia dei Lanzi (FLIR T1020). Top: RGB, open-vocabulary material segmentation (Perseus identified as bronze), and the correction map (M − B0). Bottom: constant-emissivity (B0) and material-emissivity (M) temperature, and the assigned emissivity field. The correction concentrates on the bronze; the surrounding high-emissivity architecture is essentially unchanged.*

### 4.5 Null regime and contraindications

The pulpit dataset confirms the null regime. The site is near-isothermal and predominantly high-emissivity: low-emissivity materials cover 0.23% of pixels, only three of 207 frames show a 95th-percentile correction above 0.1 K, and the whole-dataset correction magnitude is 0.033 K at the 95th percentile. ThermoScenes gives a 95th-percentile correction of 0.26 K on a high-emissivity winter facade with no low-ε materials. Across the four datasets the observed magnitudes match the regime map: negligible on the isothermal pulpit, small on the mild-contrast facades, large and directional on the high-contrast ULB17-VT scenes, and small but material-directed on the weathered-heritage Loggia. Three regimes contraindicate correction. In the isothermal regime the correction is unobservable and a small net harm under noise. On low-emissivity surfaces at high contrast the propagated uncertainty is comparable to the correction through the $1/\varepsilon$ sensitivity. And anomalies that are themselves defined by emissivity contrast, such as modern repairs, contamination, and drying fronts, are suppressed by the correction, which attributes their signature to a material emissivity difference. This motivates delivering the emissivity map and the corrected temperature map as separate, independently inspectable channels, so that a feature removed from the temperature channel as an emissivity artifact remains visible in the emissivity channel.

## 5. Discussion

The results support a bounded claim. Material-informed per-pixel emissivity improves single-band retrieval when the scene has thermal contrast, contains materials whose emissivity departs from the default, is well registered, and is well calibrated; outside that regime the correction is at best inert and can do harm. We regard the quantified regime map as the primary contribution, because it converts a practitioner intuition into an operating specification and reconciles the near-zero pulpit and small Loggia corrections as predictions of the map rather than evidence against the method. The measurement-backed audit adds a specifically heritage conclusion, that weathered heritage surfaces are high-emissivity and close to the conventional default, so the correction there is small, which is a defensible and useful position for conservators deciding when the extra RGB channel is worth acquiring. Several limitations bound the evidence. The accuracy claim rests on a synthetic benchmark; although physics-exact, using real layouts and calibration constants and breaking the inverse crime, it remains a simulation, and a laboratory capture with contact thermocouples and reference-emissivity patches is the one experiment that would test both the segmentation and the correction against physical truth on real hardware. The real datasets validate behaviour and direction but, lacking contact truth, cannot establish absolute accuracy. Segmentation is the practical limiter, and a formal multi-annotator intersection-over-union and emissivity-weighted confusion study across the quality range remains future work; open-vocabulary segmentation matches appearance rather than material identity, which motivates a self-supervised patch-feature refinement, proposed but not evaluated[33]. The reflected temperature is treated as a scalar, whereas the low-emissivity materials for which it matters most see a spatially varying field; the named-stone proxies need direct measurement; and the pipeline is not fully hands-off, since the operator must name each material as a prompt and each material needs a lookup-table entry.

## 6. Conclusion

We have presented and evaluated a training-free pipeline that derives per-pixel emissivity from open-vocabulary segmentation of a colocated RGB camera, applies it in a complete raw-digital-number inverse-Planck retrieval, and demonstrated it in situ on the Loggia dei Lanzi. Against exact synthetic ground truth the method reduces mean absolute error from 1.97 K to 0.91 K at 20 K contrast and beats the best fitted constant on every layout when the table is accurate and the palette contains low-emissivity materials; on real high-contrast data the correction follows each material emissivity deviation and leaves near-blackbody materials essentially unchanged; and on a near-isothermal heritage site it is negligible. The central deliverable is a quantified operating-regime map that specifies, in terms of thermal contrast, emissivity diversity, and lookup-table accuracy, where the correction is worth applying and where it is contraindicated. A measurement-backed audit shows that weathered heritage surfaces are high-emissivity, so the correction is modest and concentrated on genuine low-emissivity exceptions. The maps are produced as two coregistered rasters; querying them independently inside a heritage digital twin is future work we intend, not a result reported here. The immediate path to a closed absolute-accuracy claim is a standards-based laboratory capture with contact ground truth.

## AI Use Statement

During the preparation of this manuscript, the authors used Claude (Claude Opus 4.8, Anthropic[34]) for advanced literature review, parsing and consolidation of datasets, English-to-Chinese translation, and initial writing review. The authors have reviewed and edited all output and take full responsibility for the content of this publication.

## Acknowledgements

This research was funded, in part, by the US Army Corps of Engineers under research Cooperative Agreement W912HZ-17-2-0024.

## Notes

1. Gillespie, Alan, Rokugawa, Shuichi, Matsunaga, Tsuneo, et al., "A temperature and emissivity separation algorithm for Advanced Spaceborne Thermal Emission and Reflection Radiometer (ASTER) images," *IEEE Transactions on Geoscience and Remote Sensing*, vol. 36, no. 4, 1998, pp. 1113-1126.

2. Ding, De-hong, Fang, Kui, Yu, He-xiang, et al., "Multi-Emissivity Setting in Thermal Imaging Based on Visible-Light Image Segmentation," *TELKOMNIKA Indonesian Journal of Electrical Engineering*, vol. 12, no. 1, 2014, pp. 245-253.

3. Gao, Yunlai, Tian, Gui Yun, "Emissivity correction using spectrum correlation of infrared and visible images," *Sensors and Actuators A: Physical*, vol. 270, 2018, pp. 8-17.

4. Schmid, Simon, Seitz, Lukas, Menéndez Orellana, Ana E., et al., "Correction of emissivity in thermograms with neural networks," *Quantitative InfraRed Thermography Journal*, vol. 22, no. 5, 2025, pp. 519-537.

5. Bao, Fanglin, Wang, Xueji, Sureshbabu, Shree Hari, et al., "Heat-assisted detection and ranging," *Nature*, vol. 619, no. 7971, 2023, pp. 743-748.

6. Avdelidis, N. P., Moropoulou, A., "Emissivity considerations in building thermography," *Energy and Buildings*, vol. 35, no. 7, 2003, pp. 663-667.

7. Grinzato, E., Bison, P. G., Marinetti, S., "Monitoring of ancient buildings by the thermal method," *Journal of Cultural Heritage*, vol. 3, no. 1, 2002, pp. 21-29.

8. Adán, Antonio, Pérez, Víctor, Ramón, Amanda, et al., "Correction of Temperature from Infrared Cameras for More Precise As-Is 3D Thermal Models of Buildings," *Applied Sciences*, vol. 13, no. 11, 2023, pp. 6779.

9. Dang, Xinyuan, Liu, Wanqin, Hong, Qingyuan, et al., "Digital twin applications on cultural world heritage sites in China: A state-of-the-art overview," *Journal of Cultural Heritage*, vol. 64, 2023, pp. 228-243.

10. Mangano, Daniel, McAvoy, Scott, Bent, George, et al., "Protagonists of their own Discovery: Engaging the European Public with Florentine Heritage," *International Archives of the Photogrammetry, Remote Sensing and Spatial Information Sciences*, vol. XLVIII-M-9-2025, Copernicus Publications, 2025, pp. 981-988.

11. Tattersall, Glenn J., "Thermimage: Thermal Image Analysis," *R package archived on Zenodo (concept DOI, resolves to latest version; also on CRAN)*, 2017. https://doi.org/10.5281/zenodo.1069704 (accessed 2026-07-11).

12. ASTM International, "ASTM E1862-14(2018): Standard Practice for Measuring and Compensating for Reflected Temperature Using Infrared Imaging Radiometers," *ASTM International, West Conshohocken, PA, USA*, 2018.

13. McAvoy, Scott, Klingspon, Jonathan, Tong, Adrian, et al., "3D Radiometric Thermography Mosaics with Low-Cost Mobile Sensor Stack," *Remote Sensing*, vol. 18, no. 9, MDPI, 2026, pp. 1335.

14. Carion, Nicolas, Gustafson, Laura, Hu, Yuan-Ting, et al., "SAM 3: Segment Anything with Concepts," *arXiv preprint arXiv:2511.16719*, 2025.

15. Kirillov, Alexander, Mintun, Eric, Ravi, Nikhila, et al., "Segment Anything," *Proceedings of the IEEE/CVF International Conference on Computer Vision (ICCV)*, 2023, pp. 4015-4026.

16. Mineo, Simone, Pappalardo, Giovanna, "Rock Emissivity Measurement for Infrared Thermography Engineering Geological Applications," *Applied Sciences*, vol. 11, no. 9, 2021, pp. 3773.

17. Salisbury, John W., D'Aria, Dana M., "Emissivity of terrestrial materials in the 8–14 µm atmospheric window," *Remote Sensing of Environment*, vol. 42, no. 2, 1992, pp. 83-106.

18. Teodorescu, George, "Radiative Emissivity of Metals and Oxidized Metals at High Temperature," 2007. https://etd.auburn.edu/handle/10415/1395 (accessed 2026-07-11).

19. Xu, Yanfen, Zhang, Kaihua, Tian, Zhuangtao, et al., "Comparison research on spectral emissivity of three copper alloys during oxidation," *Infrared Physics & Technology*, vol. 126, 2022, pp. 104344.

20. Setién-Fernández, I., Echániz, T., González-Fernández, L., et al., "Spectral emissivity of copper and nickel in the mid-infrared range between 250 and 900\\,\textdegreeC," *International Journal of Heat and Mass Transfer*, vol. 71, 2014, pp. 549-554.

21. Park, Junghyun, Kim, Donghyun, Kim, Hyunsik, et al., "Thermal Radiative Copper Oxide Layer for Enhancing Heat Dissipation of Metal Surface," *Nanomaterials*, vol. 11, no. 11, 2021, pp. 2819.

22. Orazi, Noemi, "The study of artistic bronzes by infrared thermography: A review," *Journal of Cultural Heritage*, vol. 42, 2020, pp. 280-289.

23. Mercuri, F., Cicero, C., Orazi, N., et al., "Infrared Thermography Applied to the Study of Cultural Heritage," *International Journal of Thermophysics*, vol. 36, no. 5--6, 2015, pp. 1189-1194.

24. Zhang, Ren-Hua, Su, Hong-Bo, Tian, Jing, et al., "Non-Contact Measurement of the Spectral Emissivity through Active/Passive Synergy of CO2 Laser at 10.6 μm and 102F FTIR (Fourier Transform Infrared) Spectrometer," *Sensors*, vol. 16, no. 7, 2016, pp. 970.

25. Barreira, Eva, Almeida, Ricardo M. S. F., Simões, Maria L., "Emissivity of Building Materials for Infrared Measurements," *Sensors*, vol. 21, no. 6, 2021, pp. 1961.

26. López, G., Basterra, L. A., Acuña, L., et al., "Determination of the Emissivity of Wood for Inspection by Infrared Thermography," *Journal of Nondestructive Evaluation*, vol. 32, no. 2, 2013, pp. 172-176.

27. Steketee, J., "Spectral emissivity of skin and pericardium," *Physics in Medicine and Biology*, vol. 18, no. 5, 1973, pp. 686-694.

28. Zhang, Zhengyou, "A Flexible New Technique for Camera Calibration," *IEEE Transactions on Pattern Analysis and Machine Intelligence*, vol. 22, no. 11, 2000, pp. 1330-1334.

29. Budzier, Helmut, Gerlach, Gerald, "Calibration of uncooled thermal infrared cameras," *Journal of Sensors and Sensor Systems*, vol. 4, 2015, pp. 187-197.

30. Almasri, Feras, Debeir, Olivier, "Multimodal Sensor Fusion in Single Thermal Image Super-Resolution," *Computer Vision – ACCV 2018 Workshops*, vol. 11367, Springer, 2019, pp. 418-433.

31. Hassan, Mariam, Forest, Florent, Fink, Olga, et al., "ThermoNeRF: A Multimodal Neural Radiance Field for Joint RGB-Thermal Novel View Synthesis of Building Facades," *Advanced Engineering Informatics*, vol. 65, Elsevier, 2025, pp. 103345.

32. McAvoy, Scott Patrick, Agarwal, Aviral, Klingspon, Jonathan, et al., "Florence Thermography and XR, Dec 1st 2025 – Jan 5th 2026, Florence, Italy," *University of California, San Diego, Field Reports*, 2026. https://escholarship.org/uc/item/4tx7832v (accessed 2026-07-11).

33. Oquab, Maxime, Darcet, Timothée, Moutakanni, Théo, et al., "DINOv2: Learning Robust Visual Features without Supervision," *Transactions on Machine Learning Research*, 2024. https://openreview.net/forum?id=a68SUt6zFt (accessed 2026-07-11).

34. Anthropic, "Claude Opus 4.8 (large language model)," 2026. https://www.anthropic.com/claude (accessed 2026-07-11).